\documentclass[runningheads]{llncs}
\usepackage[utf8]{inputenc} 
\usepackage{graphicx}
\usepackage{xcolor}

\begin{document}
\title{Monte-Carlo Sampling applied to Multiple Instance Learning for Histological Image Classification}
%
%

\author{Marc Combalia\orcidID{0000-0001-5237-4256} \and
Ver\'{o}nica Vilaplana\orcidID{0000-0001-6924-9961}}
\authorrunning{M. Combalia and V. Vilaplana}

\institute{Universitat Polit\`{e}cnica de Catalunya, Barcelona, Spain\\
\email{marc.combalia@alu-etsetb.upc.edu, veronica.vilaplana@upc.edu}}

%
\maketitle              
\begin{abstract}
We propose a patch sampling strategy based on a sequential Monte-Carlo method for high resolution image classification in the context of Multiple Instance Learning. When compared with grid sampling and uniform sampling techniques, it achieves higher generalization performance. We validate the strategy on two artificial datasets and two histological datasets for breast cancer and sun exposure classification. 

\end{abstract}

\section{Introduction}
Deep learning is widely used for image classification with great success \cite{resnet} \cite{vgg}. However, neural networks can not be directly applied to very high resolution images, such as Whole Slide Tissue images, due to the high computational cost involved. A common solution consists in dividing the image into patches and using patch-level annotations to train a supervised classifier. However, patch-level annotations are not usually available, especially when working with medical datasets. On the contrary, image-level annotations are much easier to obtain so practitioners have used Multiple Instance Learning (MIL) to train patch-level classifiers in a weakly supervised manner, aggregating patch-level predictions into image-level scores \cite{bib:2017arXiv171200310T} \cite{hou2016patch} \cite{bib:milmultiinstance} \cite{attentionmil}.

When the input images are small enough, the MIL formulation can be implemented using a global Max Pooling layer at the output of a Fully Convolutional Network \cite{fcnmil}. However, in the case of high resolution images, this implementation is not possible due to memory constraints. This is why patches are usually sampled using a regular grid (with or without overlap) \cite{hou2016patch} \cite{terabyte} before being fed to the neural network. Grid sampling may skip some zones in the image which might be relevant for classification, and concentrate too much effort in zones which are not. In this work we propose a novel patch sampling strategy which extracts knowledge from the network to focus attention on the most discriminative regions in an image for a given instant in the training process, permitting better convergence and higher generalization performance. We compare this approach to uniform sampling and conventional grid sampling on two artificial and two histological datasets.


\section{Materials and methods}
\label{sec:method}

\subsection{Multiple Instance Learning}
\label{subsec:mil}

The Multiple Instance Learning formulation permits training a patch-based classifier with only image-level annotations, aggregating patch level predictions into image-level scores. 

Multiple Instance Learning is a type of weakly supervised learning algorithm where training data is arranged in bags, where each bag contains a set of instances $X=\{x_1, x_2, ..., x_M\}$, and there is one single label $Y$ per bag, $Y\in \{0,1\}$ in the case of a binary classification problem. It is assumed that individual labels $y_1, y_2, ..., y_M$ exist for the instances within a bag, but they are unknown during training. In the standard Multiple Instance assumption (SMI), a bag is considered negative if all its instances are negative. On the other hand, a bag is positive, if at least one instance in the bag is positive \cite{2016arXiv161002501W}.

The MIL formulation has been often used to solve the problem of high resolution image classification. An image (bag) is divided into $M$ patches (instances), and the patches pertaining to the same image are treated jointly in the classifier. If an image is positive ($Y = 1$), it will contain at least one positive patch ($y_{m} = 1$ at least for one m). On the contrary, if the image is negative ($Y = 0$), all its patches will be negative ($y_{m} = 0$ for all m). Then, the $max$ operator can be used to aggregate patch predictions to obtain an image-wise score: $\hat{Y} = max_m(\hat{y}_{m})$, where $\hat{y}_m$ is the prediction for patch $m$. When using this aggregating function, the weights of the network will be updated with the information of only one patch per image. Other less strict aggregating functions have been proposed in the literature \cite{bib:2017arXiv171200310T}  \cite{attentionmil} \cite{2016arXiv161002501W}, which use not only the highest scoring patch but an aggregation of more than one patch prediction per image. 


\subsection{Patch sampling}

Since the neural network will use a small subset of patches to update its weights at every iteration, it is important to select an adequate sampling strategy. The traditional grid-sampling strategy, a sampling strategy based on a uniform random variable, and a novel sampling strategy based on sequential Monte Carlo methods are reviewed in this section.

\subsubsection{Grid sampling:}

The extraction of patches is performed in a grid-like manner; the image is divided into a regular grid of patches, with or without overlap. 
Given that the sampling is performed only once in the whole training process, grid sampling is the fastest sampling strategy. However, the subset of patches to train will be the same throughout the epochs.  Also, this strategy will sample patches from not discriminative regions in the image even after the network has learned that they are not relevant for classification.


\subsubsection{Uniform sampling:}

A uniform distribution is applied to select the patches used to train the neural network at every epoch. This means that the neural network will see a different subset of patches every time the image goes through the training loop. 
As the number of training epochs increases, the network will tend to see all possible patches from every image.
Nevertheless, this approach will sample patches from irrelevant regions in the image even after they have been learned to be non-discriminative.

\subsubsection{Monte-Carlo sampling:}

The objective of this sampling strategy is to concentrate the effort on the most relevant regions of a high resolution image. When a new image is fed into the network, its output probability map is estimated using a variation of a sequential Monte-Carlo method. New patches are obtained from regions around high activations in the output probability map. 
\begin{enumerate}
\item Initialization: $n$ image points are sampled following a uniform distribution. 
\item Evaluation: a patch centered on each point is sampled and forwarded through the network. The output produced by the patch is used to represent the point.
\item Normalization: the point scores are re-scaled between 0 to 1. The points whose value is closer to 1 will be the ones corresponding to patches which have obtained the highest output from the neural network.
\item Re-sampling: the lowest scoring points are removed, and (the same number of) new points are re-sampled on top of the ones which have a higher score. This re-sampling step can be done deterministically (re-sampling the $l$ lowest scoring points) or stochastically (using a random uniform distribution).
\item Displacement: the new points are slightly displaced according to a random 2D Gaussian distribution.
\item Go to step 2 for k iterations
\end{enumerate}

The proposed method relocates patches which have not been relevant for classification into more discriminative regions in the image, that is, around the patches with higher activations. This process is performed at every batch, since the discriminative regions in the image will vary as the network learns during the training process.

\section{Experiments}
\label{sec:materials}

\subsection{Datasets}
\subsubsection{Artificial datasets:}

Two binary artificial datasets named MNIST-Sparse and MNIST-Clustered have been created to test the performance of the various sampling algorithms. Each image of 1024x1024 pixels consists of an aggregation of 28x28 images from the MNIST dataset. A positive image contains at least one MNIST digit corresponding to the class '9', while a negative image contains only digits corresponding to the other classes ('0' to '8'). Digit '9' has been chosen because it can be mistaken for '4' or '5' \cite{attentionmil}.

The purpose of these datasets is to imitate two different distributions found in histological images. The MNIST-Sparse dataset has the relevant regions (where the target number is localized) spread through the image. On the contrary, the MNIST-Clustered dataset has the target patches concentrated in space. The training and test subsets contain 1000 and 400 images, respectively. Figure \ref{fig:mnist_comparison} shows examples extracted from the two datasets, where the target digits '9' have been highlighted using red squares.

\begin{figure}[!htb]
\centering
\minipage{0.8\textwidth}
  \minipage{0.48\textwidth}
    \includegraphics[width=\linewidth]{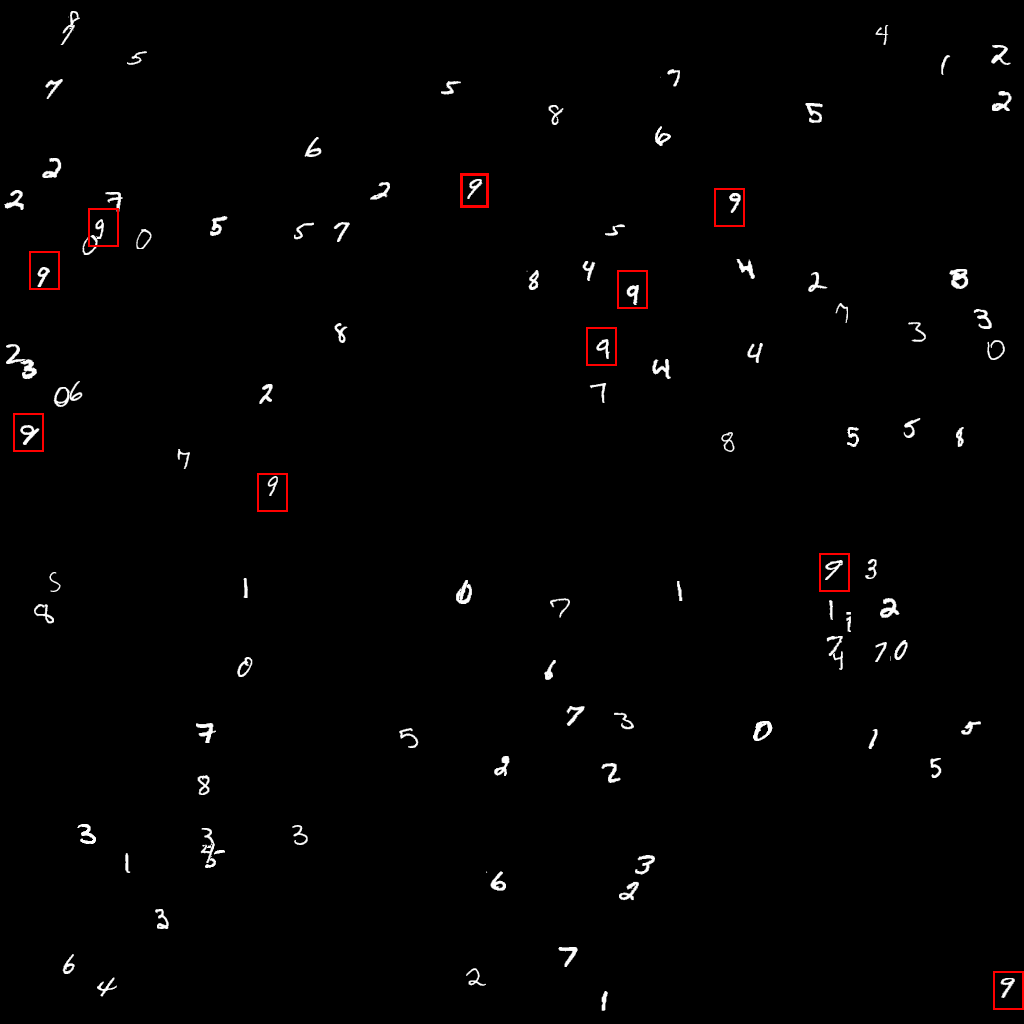}
    \centering
    \textbf{(a)}
  \endminipage
  \hfill
  \minipage{0.48\textwidth}	
    \includegraphics[width=\linewidth]{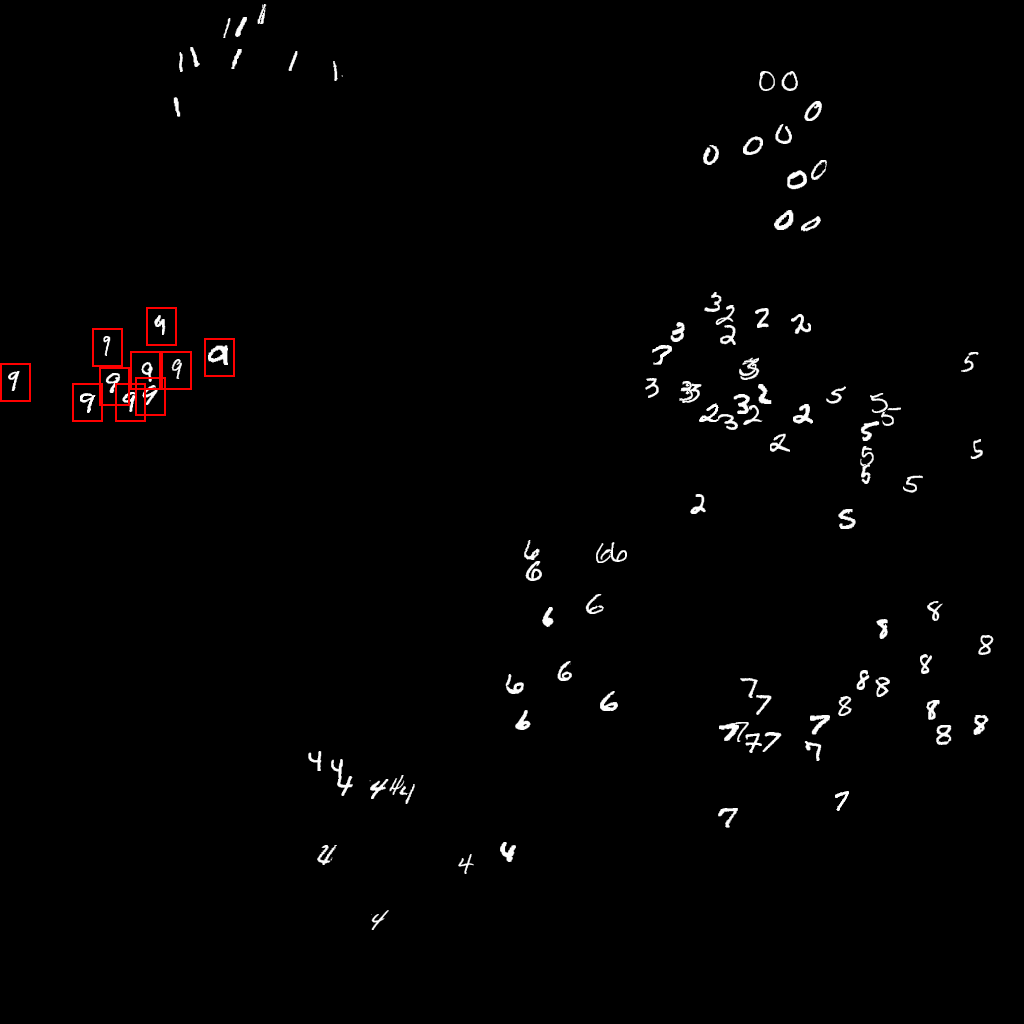}
    \centering
    \textbf{(b)}
  \endminipage
\endminipage
\caption{Images extracted from MNIST-Sparse dataset (a) and MNIST-Clustered dataset (b). Target numbers (9) are marked with red squares in both images}
\label{fig:mnist_comparison}
\end{figure}

\subsubsection{Histological datasets:}

The sampling strategies are also evaluated on two histological datasets: the ICIAR Grand Challenge 2018 dataset Part A \cite{iciar}, and the Skin subset of the GTEx dataset \cite{gtexgood}.

The ICIAR dataset is formed by 400 breast microscopy tissue images divided in 4 different classes: normal, benign, invasive and in-situ carcinoma. For each class, there are 100 different Hematoxylin \& Eosin stained images with a dimension of 2048 x 1536 pixels. The images are in RGB color space. We use benign and invasive classes to create a binary problem on which to test the algorithms, which results in 160 images for training and 40 for test. The images have already been pre-cropped from labeled Whole Slide Tissue images, and hence, from a MIL perspective, a large number of patches (instances) that we extract from an image (a bag) are expected to be consistent with the image label.

The Skin subset of the GTEx dataset is composed of approximately 10000 pieces of skin, which correspond to sun-exposed and not sun-exposed tissues. The smallest slide available (8 microns per pixel) is used to train the neural networks. The training and test splits contain 8000 and 2000 images, respectively.

\begin{figure}[!htb]
\centering
\minipage{0.23\textwidth}
  \includegraphics[width=\linewidth]{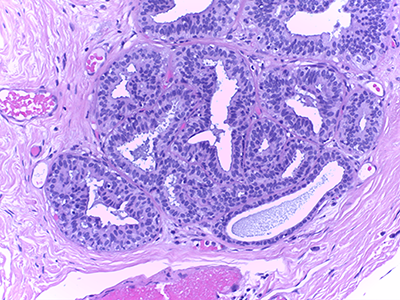}
  \centering
  \textbf{(a)}
\endminipage\hfill
\minipage{0.23\textwidth}
  \includegraphics[width=\linewidth]{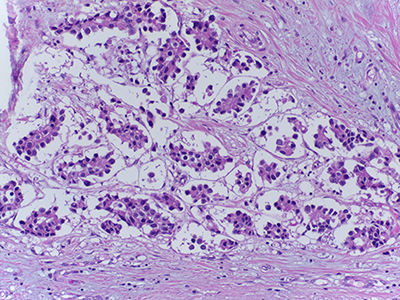}
  \centering
  \textbf{(b)}
\endminipage\hfill
\minipage{0.23\textwidth}%
  \includegraphics[width=\linewidth]{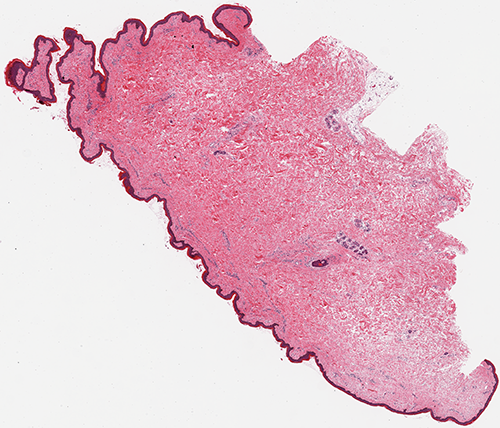}
      \centering
  \textbf{(c)}
\endminipage\hfill
\minipage{0.23\textwidth}%
  \includegraphics[width=\linewidth]{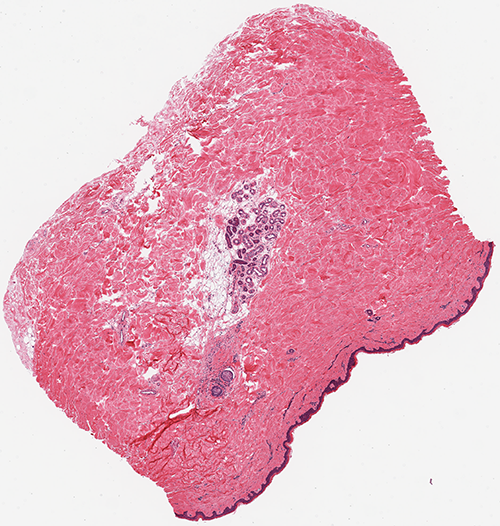}
  \centering
  \textbf{(d)}
\endminipage\hfill
\caption{Images extracted from the ICIAR Part A Dataset for benign (a) and invasive (b) classes; and the GTEx Skin dataset for the sun-exposed (c) and not-sun-exposed (d) classes.}
\label{fig:mnist_comparison}
\end{figure}

\subsection{Results}
\label{sec:experiments}

A VGG-like architecture \cite{vgg} with a receptive field of 40x40 pixels is trained to evaluate the performance of the sampling algorithms on the MNIST-Sparse and MNIST-Clustered datasets. A higher capacity neural network based on the ResNet \cite{resnet} architecture is used in the ICIAR Part A and GTEx Skin datasets with a receptive field of 224x224, since these datasets are more challenging than the MNIST toy example.

Patches are extracted without overlap for the grid sampling strategy, and the same number of patches/points is used for the uniform and Monte-Carlo training strategies. This results in a total of 625 patches per image for the MNIST dataset and 54 patches per image for the ICIAR dataset. The number of patches in the case of the GTEx dataset is variable since images have different sizes. One iteration is used on the Monte-Carlo algorithm  every time the images go through the training loop, as it was found to be enough to perform a correct estimation of the output probability map of an image.

Patch scores are aggregated using the $max$ operator into an image-level score for the MNIST and the GTEx Skin datasets, since relevant information is expected to be very localized in space. On the other hand, the Top-K (with K = 10) aggregating function is used in the ICIAR Part A dataset, since patches are expected to be consistent with the label of the image. The Top-K function will use the top K scoring instances in a bag to obtain the bag-level prediction. 

\begin{figure}[!htb]
\minipage{0.32\textwidth}
  \includegraphics[width=\linewidth]{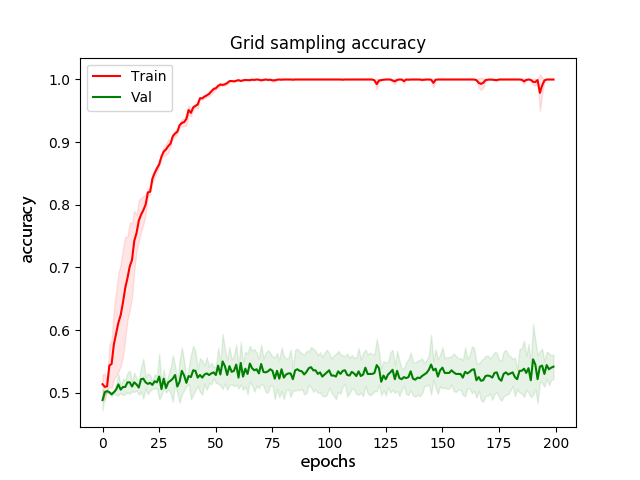}
\endminipage\hfill
\minipage{0.32\textwidth}
  \includegraphics[width=\linewidth]{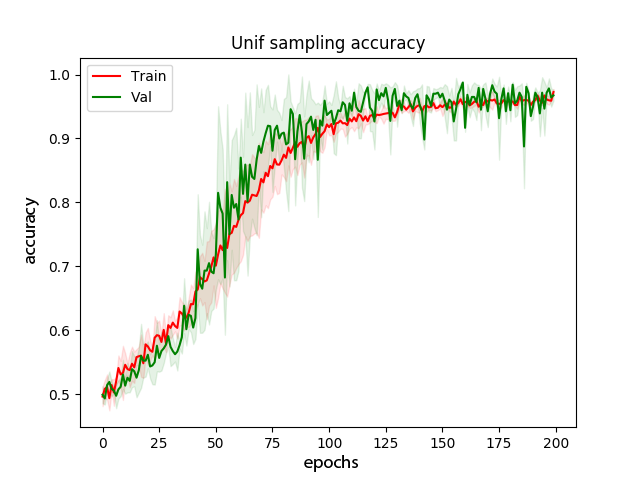}
\endminipage\hfill
\minipage{0.32\textwidth}%
  \includegraphics[width=\linewidth]{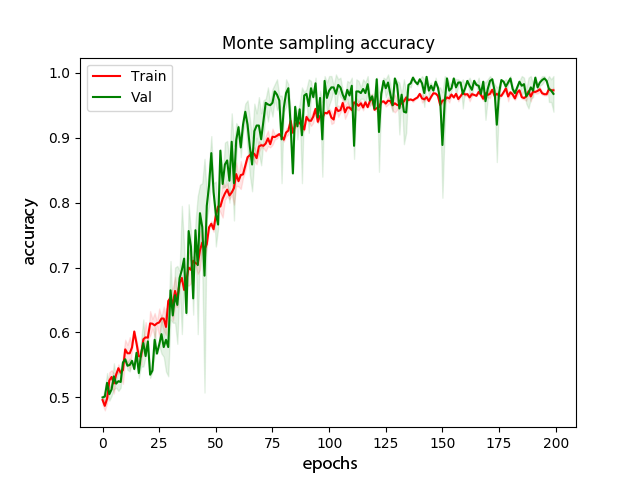}
\endminipage
\caption{Train and validation accuracy on the MNIST-Sparse dataset for the proposed sampling strategies: grid (left), uniform (center), Monte-Carlo (right).}
\label{fig:sparse_comparison}
\end{figure}

\begin{figure}[!htb]
\minipage{0.32\textwidth}
  \includegraphics[width=\linewidth]{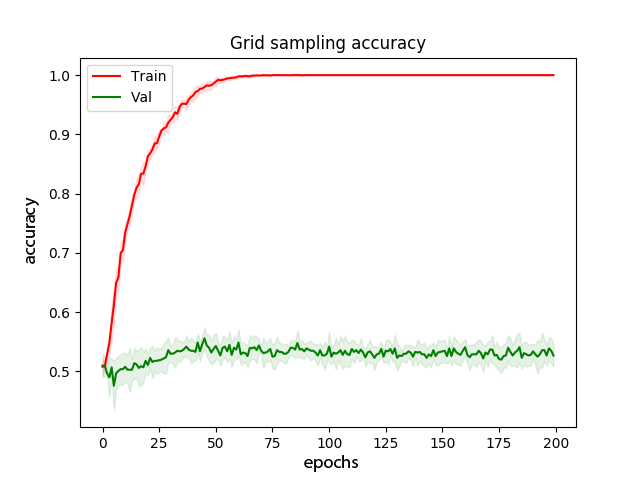}
\endminipage\hfill
\minipage{0.32\textwidth}
  \includegraphics[width=\linewidth]{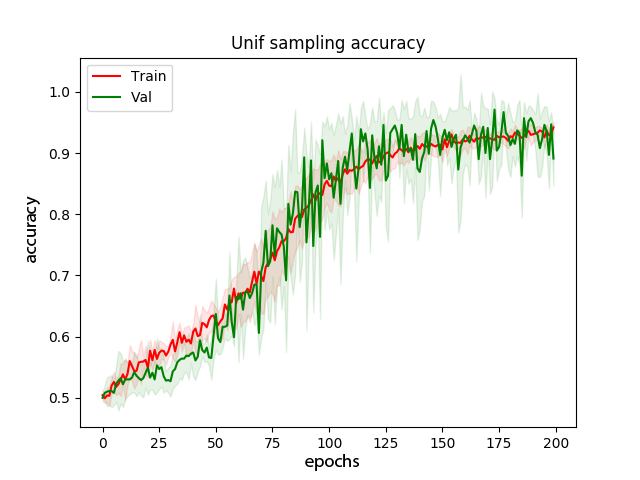}
\endminipage\hfill
\minipage{0.32\textwidth}%
  \includegraphics[width=\linewidth]{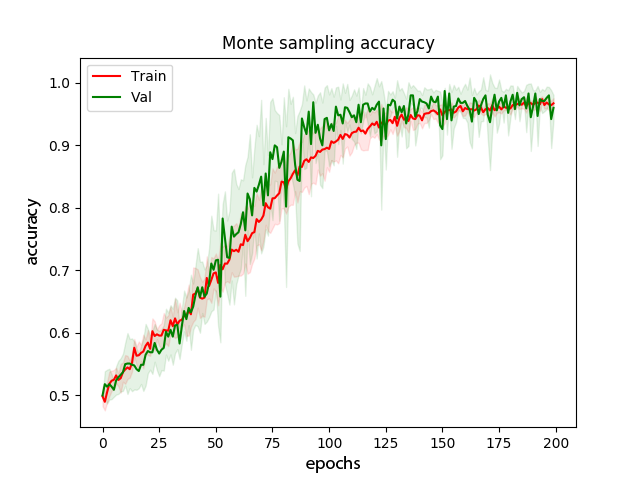}
\endminipage
\caption{Train and validation accuracy on the MNIST-Clustered dataset for the proposed sampling strategies: grid (left), uniform (center), Monte-Carlo (right).}
\label{fig:clustered_comparison}
\end{figure}

The neural networks are trained with Adam optimization. Grid sampling with 50 \% overlap is used to sample patches from the images at test time, and the max function is used to aggregate patch scores into image-wise predictions. The accuracy results on the test set for each sampling strategy are shown in Table~\ref{sample-table} and the train and validation accuracy curves for the MNIST-Sparse and MNIST-Clustered datasets are presented in Figures \ref{fig:sparse_comparison} and \ref{fig:clustered_comparison}, respectively.

\begin{table}[ht]
  \caption{Test accuracies for the various sampling strategies on the MNIST-Sparse, MNIST-Clustered, ICIAR Part A and GTEx Skin datasets}
  \label{sample-table}
  \centering
  \begin{tabular}{|c|c|c|c|c|}
    \hline
    Test accuracy & MNIST Sparse & MNIST Clust & ICIAR PartA & GTEx Skin \\
    \hline
    Grid sampling  & 0.520 $\pm$ 0.01 & 0.523 $\pm$ 0.01 & 0.776 & 0.826\\
    Uniform sampling  & 0.759 $\pm$ 0.03 & 0.83 $\pm$ 0.01  & 0.790 & 0.920 \\
    Monte-Carlo sampling & \textbf{0.825 $\pm$ 0.02} & \textbf{0.852 $\pm$ 0.03} & \textbf{0.847} & \textbf{0.942} \\
	\hline
  \end{tabular}
\end{table}

\section{Discussion}

Figures \ref{fig:sparse_comparison} and \ref{fig:clustered_comparison} illustrate how the sampling strategy used for training can have a very large impact on the final performance of the neural network.
This is especially true in cases where the receptive field of the network is small compared to the spatial extension of the discriminative features, as it is on the MNIST-Sparse and MNIST-Clustered datasets, where target digits are patches of 28x28 pixels and clusters of 40x40 pixels, respectively. In these cases, the grid sampling technique would need a very large overlap between patches to provide a good subset of patches to the network, which would result in a substantial increase in training time. While the neural network trained with the grid sampling strategy is unable to learn the true distribution of the data, the stochastic nature of the uniform and Monte-Carlo sampling strategies permits seeing a different subset of patches at every epoch. This allows the network to correctly find the discriminative regions on the image. In addition, once the network has learned which regions in the image are the relevant ones, the Monte-Carlo strategy samples only from these regions. This results in a higher validation accuracy, especially in the MNIST-Clustered dataset.

\begin{figure}[!htb]
\centering
\minipage{0.23\textwidth}
  \includegraphics[width=\linewidth]{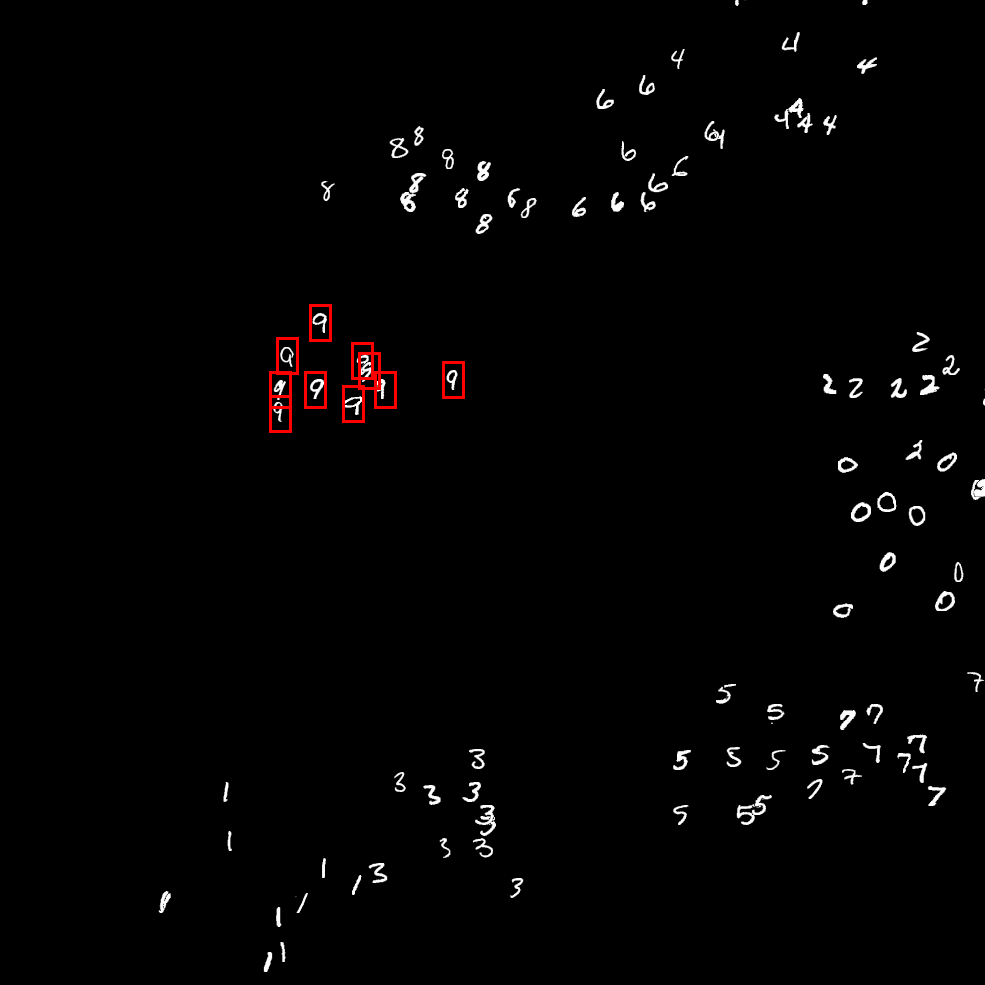}
  \centering
  \textbf{(a)}
\endminipage\hfill
\minipage{0.23\textwidth}
  \includegraphics[width=\linewidth]{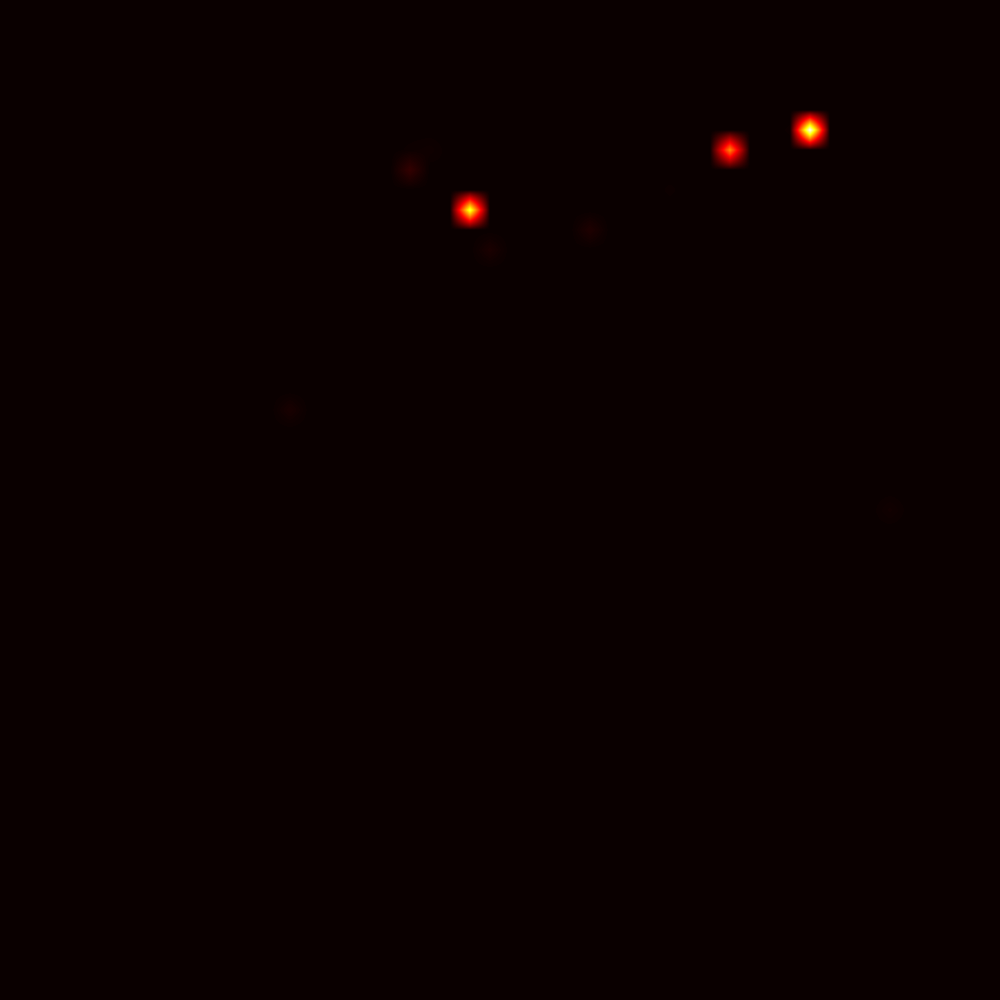}
  \centering
  \textbf{(b)}
\endminipage\hfill
\minipage{0.23\textwidth}%
  \includegraphics[width=\linewidth]{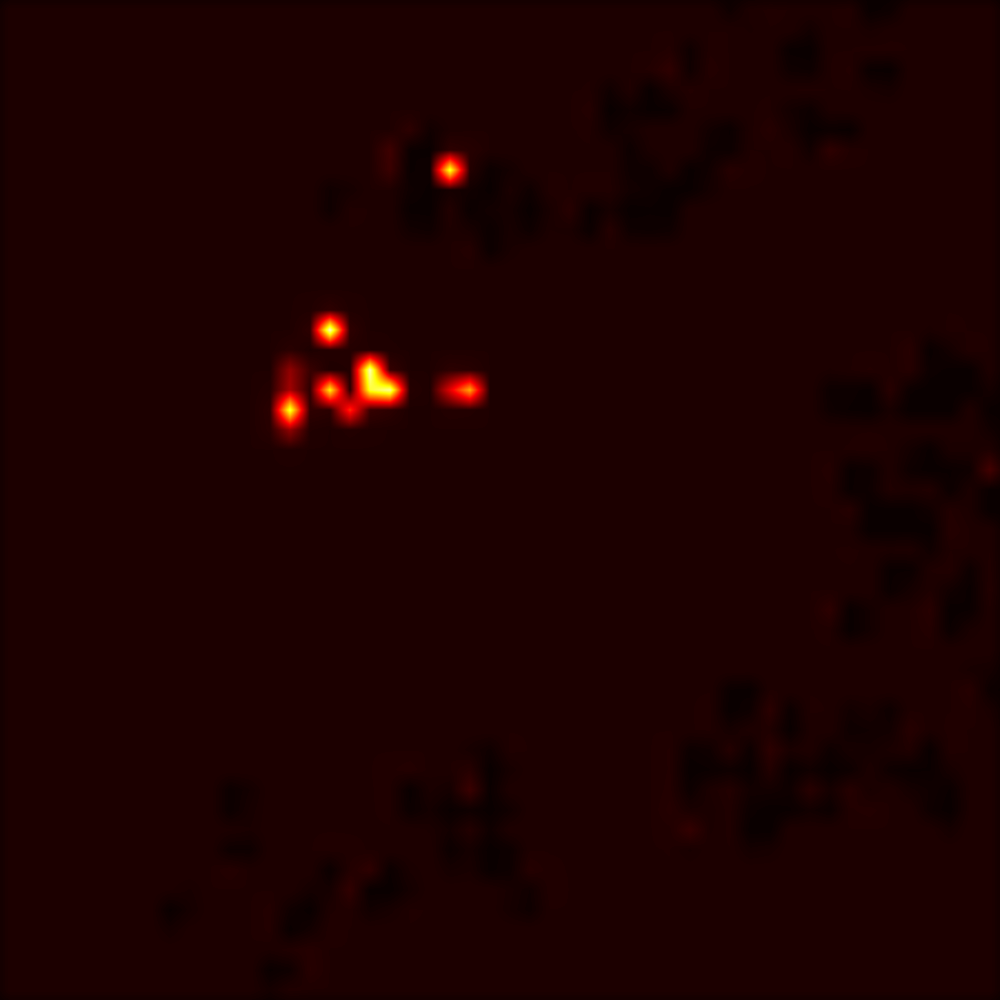}
      \centering
  \textbf{(c)}
\endminipage\hfill
\minipage{0.23\textwidth}%
  \includegraphics[width=\linewidth]{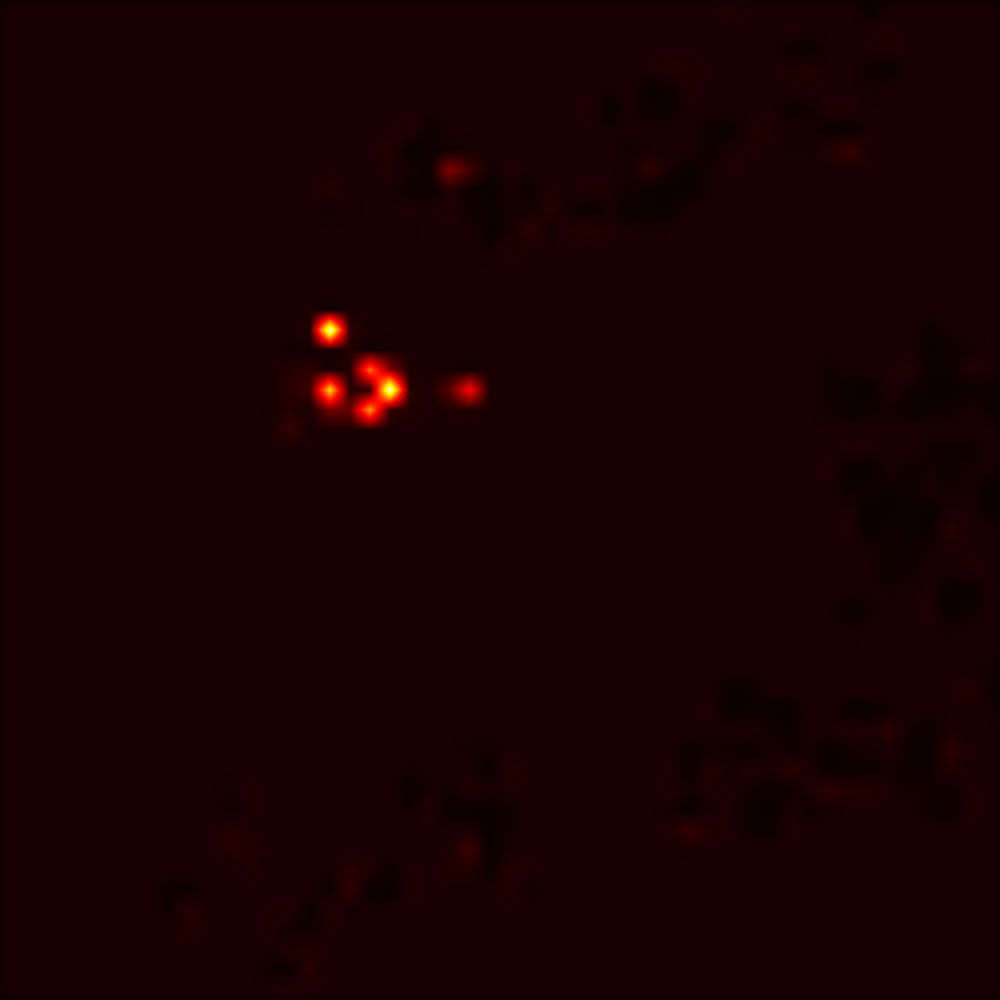}
  \centering
  \textbf{(d)}
\endminipage\hfill
\vspace{0.2cm}
\minipage{0.12\textwidth}%
  \includegraphics[width=\linewidth]{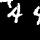}
      \centering
  \textbf{(e)}
\endminipage \hspace{1cm}
\minipage{0.12\textwidth}%
  \includegraphics[width=\linewidth]{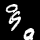}
  \centering
  \textbf{(f)}
\endminipage \hspace{1cm}
\minipage{0.12\textwidth}%
  \includegraphics[width=\linewidth]{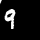}
  \centering
  \textbf{(g)}
\endminipage
\caption{Input image (a), output probability map for grid sampling (b), uniform sampling (c) and Monte-Carlo sampling (d). Figures (e), (f) and (g) show the maximally activated patch for grid, uniform and Monte-Carlo sampling, respectively.}
\label{fig:output_maps}
\end{figure}

Figure \ref{fig:output_maps} shows the output probability maps of one positive image from the MNIST-Clustered database for each sampling strategy. The neural network trained with the grid sampling strategy fails to localize the target digit; the patch with the maximum score contains a '4'. On the other hand, the neural networks trained with the stochastic sampling strategies succeed activating on the regions containing the target digit '9'. The Monte-Carlo sampling strategy produces a more accurate map.

\begin{figure}[h]
\centering
\includegraphics[width=0.8\textwidth]{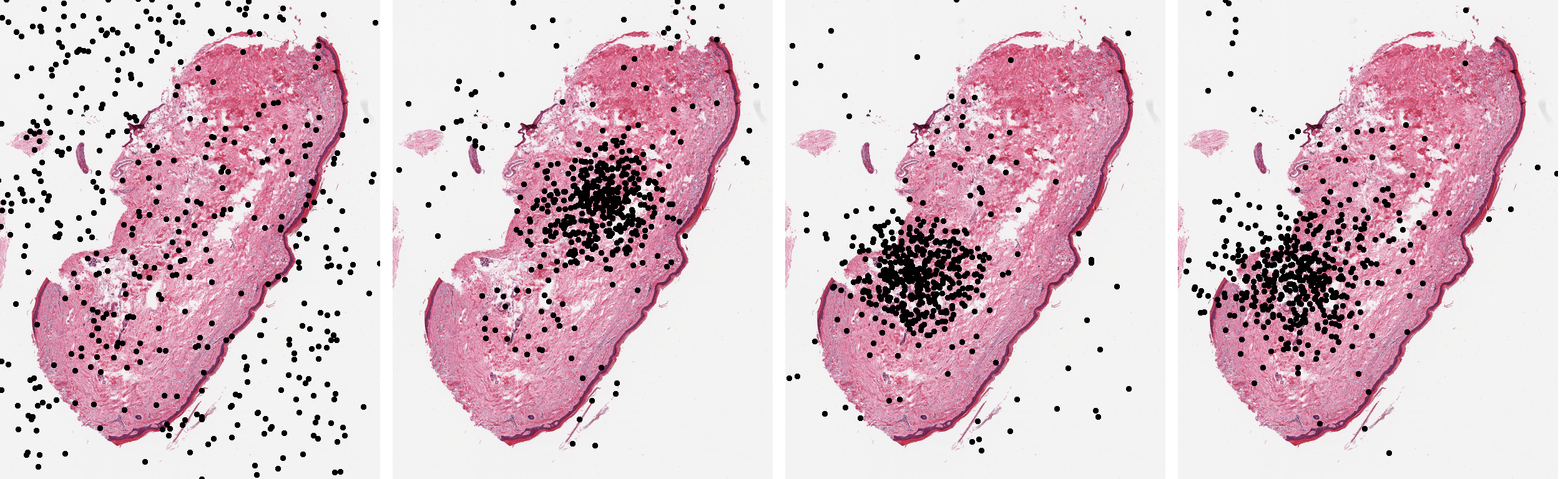}
\caption{Black points correspond to points sampled with Monte-Carlo at epochs 2, 6, 8 and 37 during the training process of the neural network for the sun-exposure classification problem.}
\label{fig:example}
\end{figure}

The stochastic sampling approaches outperform the grid sampling also on the histological datasets. This time, however, the neural networks trained with the grid sampling strategy perform correctly. In this case the neural networks have a larger receptive field, and the discriminative image regions are smaller compared to the receptive field. However, the neural networks still benefit from the Monte-Carlo training strategy, which focuses on the relevant regions, improving accuracy. Figure \ref{fig:example} shows how the distribution of samples in the Monte-Carlo approach changes as the neural network learns. In the first epochs, the Monte-Carlo strategy behaves very similarly to the uniform sampling strategy. However, as the network keeps learning, the Monte-Carlo sampling strategy further concentrates its effort on the discriminative regions.




\section{Conclusions}
\label{sec:conclusion}
In this paper we have shown that a simple grid sampling technique can compromise the performance of a network, especially when its receptive field is small compared to the size of the relevant features in the image. We have proposed a sampling strategy based on a sequential Monte-Carlo method for high resolution images which samples from the most relevant regions during the training process, overcoming the problems of grid sampling. We have illustrated its capabilities on two artificial and two histological datasets for breast cancer and sun exposure classification.

%
%
%
%
\subsubsection{Acknowledgments:} This work has been partially supported by the project MALEGRA TEC2016-75976-R financed by the Spanish Ministerio de Econom\'{i}a y Competitividad and the European Regional Development Fund.

\bibliographystyle{splncs04}
\bibliography{mybib}

\end{document}